\documentclass{article}
\usepackage[numbers,sort&compress]{natbib}
\usepackage{arxiv}
\usepackage{capt-of}
\usepackage[utf8]{inputenc} 
\usepackage[T1]{fontenc}    
\usepackage[hidelinks]{hyperref}       
\usepackage{xurl}            
\usepackage{booktabs}       
\usepackage{amsfonts}       
\usepackage{nicefrac}       
\usepackage{microtype}      
\usepackage{lipsum}
\usepackage{graphicx}
\graphicspath{ {./images/} }
\usepackage{amsmath,amsfonts,amssymb}
\usepackage{placeins}
\title{Mapping the Emergence of Regularization-Driven Dynamics in Grokking}
\date{October 6, 2026}
\hypersetup{
  pdftitle={Mapping the Emergence of Regularization-Driven Dynamics in Grokking},
  pdfauthor={Yiming Lin and Yuxuan Wang}
}

\author{Yiming Lin\\
School of Artificial Intelligence,\\
University of Chinese Academy of Sciences\\
\texttt{linyiming25@mails.ucas.ac.cn}
\And
Yuxuan Wang\\
Chinese Institute for Brain Research (cibr.ac.cn)\\
\texttt{wangyuxuan@cibr.ac.cn}}

\begin{document}
\maketitle
\begin{abstract}
For overparameterized neural networks, many solutions can fit the training data equally well while behaving very differently on unseen samples. Grokking separates training fit from visible generalization, providing a window for studying how this selection develops during training. We sweep short, fixed-duration weight-decay (WD) perturbations across the pre-generalization plateau and measure how they shift later generalization time. Across three grokking tasks, these shifts are unordered early in the plateau but later form a stable dose ordering before visible generalization, with stronger WD increases leading to earlier generalization and stronger WD decreases leading to later generalization. Test-loss barriers between perturbed and baseline generalization checkpoints collapse toward zero while the ordered timing effects persist. A similar response reorganization is observed under $\ell_1$ regularization in the grokking setting of \citet{junior2025grokking}. Drawing on Waddington's developmental landscape as an analogy, we call this combination of increasingly constrained solution selection and persistent dose-ordered timing shifts the canalization of grokking solution selection. Together, our response maps and loss-barrier measurements reveal a dynamical reorganization before visible generalization that is consistent with the theoretical picture of regularization-driven motion along a stable slow manifold \citep{boursier2025grokking}.
\end{abstract}

\section{Introduction}
\label{sec:introduction}
How neural networks learn generalizable patterns from limited training samples is a foundational question in understanding deep learning \citep{zhang2017understanding,jiang2019predicting}. For overparameterized neural networks, the training data usually allow for many solutions that fit the training set equally well but behave very differently on unseen samples. The training objective alone does not determine the final solution; which solution is reached depends on the implicit bias of the optimization process \citep{gunasekar2018characterizing}. In standard supervised learning, this process usually happens alongside fitting the training set and improving test performance, which makes it hard to study on its own. Grokking describes a form of delayed generalization. A model can first fit the training set, then remain on a long plateau of low test performance, and only generalize rapidly after further training \citep{power2022grokking,nanda2023progress}. This temporal separation between training fit and visible generalization provides a clear window for observing and perturbing the optimization process before generalization appears.

Previous work has repeatedly found that neural network training proceeds through distinct dynamical stages. In grokking, this can include a transition from lazy to feature-learning dynamics \citep{kumar2024grokking}. Studies of loss geometry and neural tangent kernel evolution suggest that training begins with a short, highly sensitive phase and later enters a more stable and constrained regime \citep{fort2020deep,zhou2025cone}. Perturbation studies have similarly found that very small parameter changes early in training can send trajectories in very different directions, while the same perturbations have much smaller effects later in training \citep{frankle2020linear,altintas2025butterfly,zhou2025twophase}.

These results suggest that the downstream effects of local perturbations can reveal how training dynamics change over time. Weight decay (WD) provides a natural intervention target for this purpose. Its effect depends strongly on when it is applied \citep{golatkar2019time}, and WD is also known to strongly influence whether and when generalization occurs in grokking \citep{power2022grokking,liu2022towards,lyu2024dichotomy,pearce2023memorize}.  Recent theoretical work connects delayed generalization to slow, regularization-driven solution selection after training fit \citep{boursier2025grokking,xu2026togrok,junior2025grokking}. Building on this picture, we ask when and how training enters a stable regime in which regularization pressure governs the pace of progress toward generalization, and how this transition differs across tasks.

Building on this, we introduce an intervention-based framework that uses the timing of generalization in grokking as a dynamical probe of how the training state responds to local perturbations over the course of optimization. We slide a fixed-duration intervention window across the entire pre-generalization plateau (after training-fit, before test accuracy exceeds chance), temporarily increasing or decreasing WD before restoring it to its baseline value and continuing training. We call this short local intervention a WD pulse. By scanning the entire pre-generalization plateau, we examine how short WD pulses applied at different stages of training affect outcomes much later in training. After the pulse ends, the model returns to the original training conditions, but the effect of the pulse can persist and later appear as earlier or later generalization. Recent work has intervened on grokking by persistently modifying training conditions \citep{sivasankar2026circuit,pandey2026thermodynamic,lyle2025nonstationarity,truong2026weightnorm}. In contrast, our WD pulses are temporary and are swept across the pre-generalization plateau to probe how the training state responds to a transient local perturbation.

We apply this framework to three algorithmic grokking tasks with clear chance-level plateaus. By scanning both when the WD pulse is applied and how strong it is, we record the resulting shift in generalization time. These measurements form a WD-response map, which reveals a dynamical reorganization before any visible generalization. Early in the plateau, generalization-time shifts show no stable ordering across pulse magnitudes. Later, the responses reorganize into a stable dose-dependent structure, with stronger positive WD pulses leading to earlier generalization and stronger negative WD pulses leading to later generalization. This structure emerges while test accuracy is still near chance level, and reproduces across random initializations and tasks. We further observe the similar pre-generalization emergence of dose-ordered timing responses in a separate L1-regularized grokking setting \citet{junior2025grokking}, indicating that this response reorganization is not specific to decoupled weight decay. Linear mode connectivity analysis \citep{frankle2020linear,entezari2022role} further shows that, as the pulse is applied later in training, test-loss barriers between the perturbed branches and the baseline at their matched generalization checkpoints collapse toward zero, while the ordered shifts in generalization time persist. 
These results suggest the emergence of a regime in which regularization governs the pace of progress toward generalization while solution selection becomes increasingly constrained. Borrowing Waddington's concept of \textit{canalization} \citep{waddington1942canalization}, we refer to this reorganization as the \textit{canalization} of grokking solution selection. These observations support interpreting the late pre-generalization dynamics as regularization-driven motion along a stable slow manifold \citep{boursier2025grokking}. Our response maps and loss-barrier measurements characterize when and how this regularization-driven regime becomes established.

\section{Grokking as a Dynamical Probe}
\label{sec:dynamical_probe}

\subsection{Experimental Tasks}

We use a simple parity-matching task (Task 1) as our main experimental setting, and adopt the sparse-parity (Task 2) setup of Merrill et al.~\citep{merrill2023tale} and the modular-addition (Task 3) setup of Google PAIR~\citep{pearce2023memorize} to test whether the same intervention pattern appears across different rule-learning tasks. In parity matching, each example consists of two integer indices, and the label indicates whether they have the same parity. Related sparse-parity and modular-addition settings have been shown to exhibit measurable internal progress before visible generalization \citep{barak2022hidden,merrill2023tale,nanda2023progress}. Task~1 provides a useful contrast; when we adapt these progress measures to parity matching, restricted loss begins to decrease only slightly before test loss (about $50$ epochs), parameter movement levels off early in training, and feature amplification changes only around the same time as test loss (see Appendix~\ref{app:progress_measures}). 

Our analysis focuses on runs that exhibit the canonical grokking pattern with a clear chance-level plateau after the training set is fit, followed by a relatively sharp rise in test performance. Full task definitions, model architectures, training configurations, and seed-selection criteria are provided in Appendix~\ref{app:tasks_and_baselines}.

\begin{figure*}[t]
    \centering
    \includegraphics[width=\textwidth]{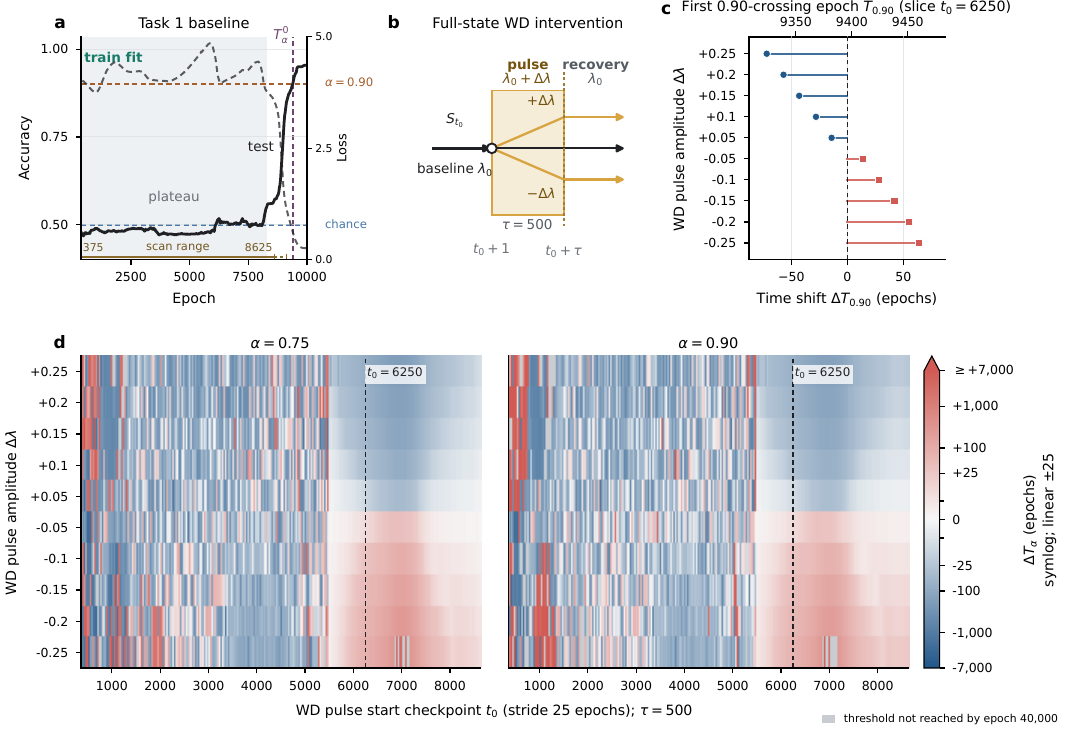}
\caption{\textbf{Weight-decay pulse interventions and construction of the WD-response map.}
An example run from the Task~1 parity-matching task is shown.
\textbf{(a)} Baseline training and test accuracy. The horizontal bar at the bottom marks the scan range of pulse start times, beginning at full training fit (epoch $375$) and ending $1{,}000$ epochs before the baseline first reaches $95\%$ test accuracy (epoch $8625$).
\textbf{(b)} For each $t_0$, all perturbed branches start from the same full training state $\mathcal{S}_{t_0}$. Weight decay is changed for $\tau$ epochs and then restored to its baseline value.
\textbf{(c)} Generalization-time responses to different WD perturbation magnitudes at a fixed pulse start $t_0$.
\textbf{(d)} WD-response maps for $\alpha=0.75$ and $\alpha=0.90$. This run contains $331$ pulse start times and $3{,}310$ perturbed branches. Early in the plateau, the responses show no stable ordering; around $t_0\approx5000$--$5500$, they begin to form a persistent dose-dependent structure, while the baseline is still on the chance-level plateau shown in panel~(a), before any visible generalization. Intervention and scan settings are given in Appendix~\ref{app:wd_implementation}.}
\label{fig:wd_response_overview}
\end{figure*}

\subsection{Sliding-Window WD Interventions}

All models are trained with AdamW. In AdamW, weight decay is decoupled from the loss-gradient update \citep{loshchilov2019decoupled}, allowing us to temporarily change the weight-decay coefficient without modifying the loss function or gradient computation.

For each baseline training trajectory, we slide a fixed-duration intervention window along the pre-generalization plateau. The scan begins at the first time the baseline reaches \(100\%\) training accuracy and then advances with stride \(\delta t\). At each intervention start \(t_0\), we branch from the full training state \(\mathcal{S}_{t_0}\) saved at that time, including the model parameters, AdamW optimizer state, random-number-generator state, and data-sampling state (Figure~\ref{fig:wd_response_overview}(b); implementation details are given in Appendix~\ref{app:wd_implementation}). We then temporarily change the baseline weight decay \(\lambda_0\) for a duration \(\tau\):

$$
\lambda(t)=
\begin{cases}
\lambda_0+\Delta\lambda,
& t_0 < t \leq t_0+\tau, \\[4pt]
\lambda_0,
& t > t_0+\tau .
\end{cases}
$$

Here, \(\Delta\lambda>0\) increases weight decay and \(\Delta\lambda<0\) decreases it. After the intervention window, training resumes with the baseline weight decay, while all other training conditions remain unchanged. Each \((t_0,\Delta\lambda)\) pair therefore defines one short WD intervention.

Across the three tasks, the main analysis includes 70 baseline runs, 23,926 intervention starts, and 239,260 intervention branches in total (Table~\ref{tab:intervention_scale}).

\subsection{Generalization-Time Response}

We use the first time a model reaches a predefined test-accuracy threshold to track its progress toward generalization. Let this threshold be $\alpha$. We denote by $T_\alpha^0$ the first time the baseline reaches $\alpha$, and by $T_\alpha(t_0,\Delta\lambda)$ the first time a perturbed branch reaches the same threshold after a WD pulse of magnitude $\Delta\lambda$ is applied at time $t_0$. We define
\[
\Delta T_\alpha(t_0,\Delta\lambda)
=
T_\alpha(t_0,\Delta\lambda)-T_\alpha^0 .
\]
Thus, $\Delta T_\alpha$measures the change in generalization time caused by the WD pulse; negative values indicate earlier generalization, while positive values indicate later generalization.

By scanning both the pulse start $t_0$ and perturbation magnitude $\Delta\lambda$, we obtain the WD-response map shown in Figure~\ref{fig:wd_response_overview}(d). Each location in the map corresponds to one WD intervention, and its response value shows how much that intervention advances or delays generalization.

\section{Results}
\label{sec:results}
\subsection{WD Responses Become Ordered Before Visible Generalization}
\label{sec:wd_response_ordering}
\begin{figure*}[t]
    \centering
    \includegraphics[width=\textwidth]{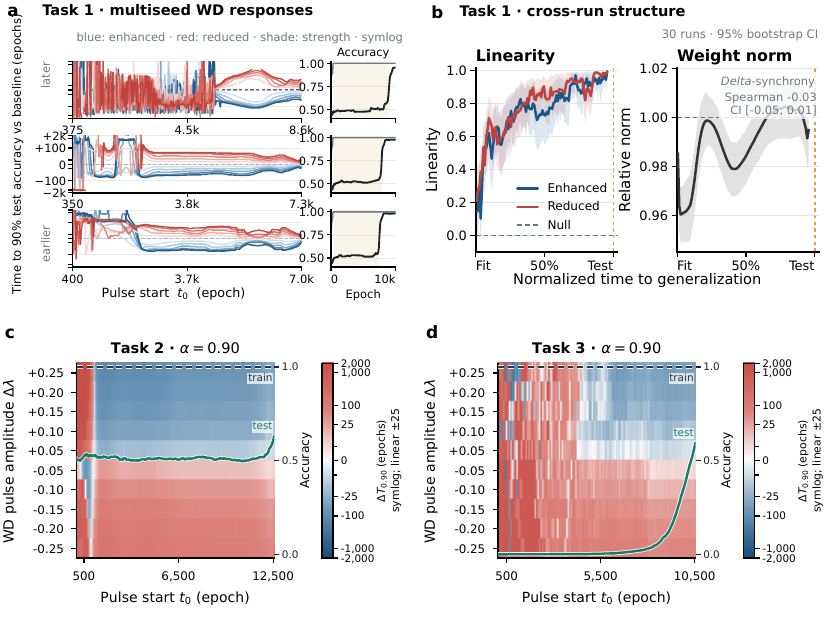}
    \caption{\textbf{WD responses develop a stable dose-dependent structure before visible generalization, and this pattern is reproduced across initializations and tasks.}
    \textbf{(a)} WD-response curves ($\alpha=0.90$) for three example Task~1 runs; insets show the corresponding baseline training and test accuracy.
    \textbf{(b)} Mean dose--response linearity $R^2_{\mathrm{corr}}$ across 30 Task~1 runs (left) and baseline relative weight norm (right).
    \textbf{(c,d)} WD-response maps ($\alpha=0.90$) for example runs from Task~2 and Task~3.
    Task and training settings are given in Appendix~\ref{app:tasks_and_baselines}; dose--response and weight-norm analyses in Appendix~\ref{app:dose_response_analysis}; and intervention settings in Appendix~\ref{app:wd_implementation}.
    Figures~\ref{fig:cross_seed_barrier_wdr_task1}--\ref{fig:cross_seed_barrier_wdr_task3} show the signed $\Delta T_{0.95}$ response maps for all 70 analyzed runs, allowing the direction, dose ordering, and temporal persistence of the responses to be inspected run by run.}
    \label{fig:wd_response_across_runs_tasks}
\end{figure*}

Figure~\ref{fig:wd_response_overview}(d) shows that the effect of WD perturbations changes over the course of training. Early in the plateau ($t_0 \lesssim 5000$), there is no stable relationship between $\Delta T_{\alpha}$ and the direction or magnitude of $\Delta\lambda$. By the middle of the plateau ($t_0 \approx 5000$--$5500$), the responses begin to form a clear and persistent dose-dependent structure; stronger positive WD perturbations lead to earlier generalization, whereas stronger negative WD perturbations lead to later generalization. Figure~\ref{fig:wd_response_across_runs_tasks}(a) shows the same transition from another perspective. Each curve corresponds to a fixed WD perturbation. The response curves cross frequently early in the plateau, but later become stably ordered by $|\Delta\lambda|$. When this stable ordering emerges, the baseline test accuracy is still at chance level, indicating that this dynamical reorganization occurs before visible generalization.

Figure~\ref{fig:wd_response_across_runs_tasks}(b) summarizes this process across 30 Task~1 runs. On a common phase axis, we separately compute the chance-corrected dose--response linearity $R^2_{\mathrm{corr}}$ for positive and negative WD perturbations (see Appendix~\ref{app:dose_response_linearity} for the definition). Both mean curves increase over training and approach $1$ near the end of the interval, indicating that the WD responses gradually develop an approximately linear dose relationship as training proceeds, and that this pattern is reproduced across different initializations.

In AdamW, weight decay contributes a term $-\eta\lambda\theta$ to the parameter update independently of the loss gradient. Its most direct effect is therefore to change the rate of parameter shrinkage and, in turn, the weight norm. We therefore examine whether the emergence of the dose-dependent response structure is synchronized with changes in the baseline weight norm. We find that the baseline relative weight norm remains nearly constant throughout the plateau in the same set of runs, staying within about $\pm 4\%$ of its value at training fit. Moreover, changes in relative weight norm and $R^2_{\mathrm{corr}}$ across phase bins show almost no synchrony. Computing the Spearman correlation between these two sets of changes for each run gives a median of $\rho_\Delta=-0.03$, with a 95\% CI of $[-0.05, 0.01]$ (see Appendix~\ref{app:linearity_norm_synchrony} for the definition; Figure~\ref{fig:wd_response_across_runs_tasks}(b), right). These results argue against slow drift in the baseline global weight norm as the explanation for the response reorganization.

Figure~\ref{fig:wd_response_across_runs_tasks}(c,d) shows the results for Task~2 and Task~3. Despite differences in their time scales and detailed response patterns, both tasks show the same qualitative behavior as Task~1. Across initializations and tasks, the generalization-time response to WD perturbations changes over training from an unordered pattern early in the plateau to a stable dose ordering later on. More importantly, this response becomes ordered before visible generalization.

\subsection{Loss Barriers Collapse While Ordered WD Responses Persist}
\label{sec:barrier_collapse}
For each intervention, we take the checkpoints at which the perturbed branch and the baseline first reach the same test accuracy and linearly interpolate between them in parameter space \citep{frankle2020linear}. The barrier is the maximum test loss along this path in excess of the linear interpolation of the endpoint losses \citep{entezari2022role}. A barrier near zero means the two checkpoints are connected by a low-loss path; a large barrier means they are linearly separated by a high-loss region. The full definition is given in Appendix~\ref{app:wd_implementation}.

\begin{figure*}[!t]
    \centering
    \includegraphics[width=\textwidth]{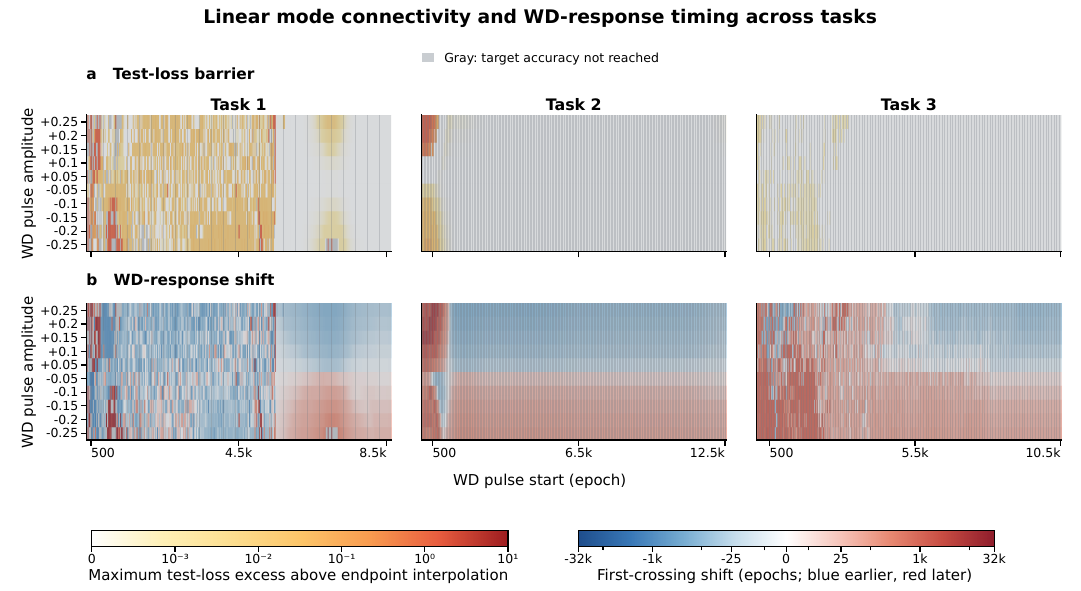}
\caption{\textbf{Test-loss barriers collapse while ordered WD timing responses persist.}
Each column shows one example run from a task (the Task~1 run is the same as in Figure~\ref{fig:wd_response_overview}), and the two rows share the same $(t_0,\Delta\lambda)$ intervention grid. For both the baseline and each perturbed branch, the generalization checkpoint is defined as the first time test accuracy reaches $\alpha=0.95$.
\textbf{(a)} Test-loss barrier between the perturbed and baseline generalization checkpoints. We evaluate test loss at $51$ equally spaced interpolation points along the straight line in parameter space connecting the two checkpoints, and define the barrier as the maximum excess above the linear interpolation of the endpoint losses (see Appendix~\ref{app:wd_implementation} for the definition). Gray cells indicate perturbed branches that do not reach the target accuracy within the training limit.
\textbf{(b)} Generalization-time shift $\Delta T_{0.95}$ of each perturbed branch relative to the baseline. Negative values indicate earlier generalization, and positive values indicate later generalization.}
\label{fig:mode_connectivity_response}
\end{figure*}
Figure~\ref{fig:mode_connectivity_response}(a) shows that, as the intervention is applied later in training, the test-loss barrier between the perturbed and baseline generalization checkpoints gradually decreases toward zero. This pattern is very clear in Task~1; the barrier drops rapidly around the onset of stable dose ordering, with stronger perturbations retaining larger barriers. Task 2 and Task 3 show the same late-stage trend, and in Task 3 the barriers are already low well before the dose ordering emerges. Overall, as the intervention is moved later, the perturbed and baseline generalization checkpoints become increasingly linearly connected by low-test-loss paths. The WD timing responses in Figure~\ref{fig:mode_connectivity_response}(b), however, retain a clear directional and dose-dependent structure. This ordered effect on generalization time persists even when the test-loss barrier is already close to zero.

Together, these results suggest that, before visible generalization, WD perturbations become increasingly less able to push the model toward generalization solutions that are linearly separated from the baseline. Later interventions mainly affect the timing of generalization rather than driving the model toward linearly separated generalization checkpoints. The collapse of the test-loss barrier and the persistence of ordered WD responses are both reproduced across random initializations (30 Task~1 runs, 24 Task~2 runs, and 16 Task~3 runs; see Appendix~\ref{app:cross_seed_analysis}).

A similar pattern appears in PCA projections of the parameter-update directions. Early pulses produce clear deviations from the baseline, whereas pulses applied when dose ordering first emerges yield trajectories that largely overlap with the baseline in the same projection (see Appendix~\ref{app:update_directions} and Figure~\ref{fig:update_direction_pca}). In these paired comparisons, the pulse starts differ by only 500 epochs, and both sets of branches are followed for several thousand epochs using a shared PCA basis within each run (see Appendix~E and Figure~8). These observations provide complementary evidence against insufficient remaining training time as the primary explanation for the barrier collapse.

\subsection{Ordered Responses Extend Beyond Weight Decay}

We further examine whether the emergence of structured perturbation responses depends specifically on decoupled weight decay. We apply the same transient-intervention framework to the $\ell_1$-regularized grokking setting introduced by \citet{junior2025grokking}. The task definition, model architecture, and baseline training configuration follow their implementation, while we introduce temporary changes to the $\ell_1$ regularization strength during training. Full details of the setting and intervention protocol are provided in Appendix~\ref{app:l1}.

\begin{figure}[t]
    \centering
    \includegraphics[width=\linewidth]{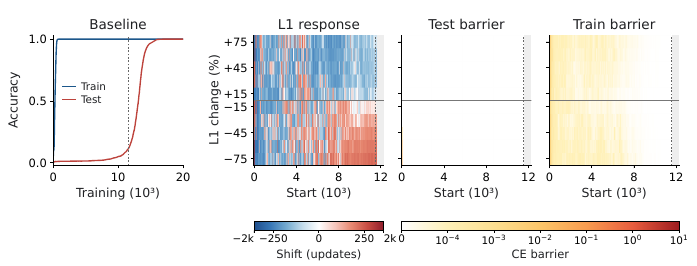}
    \caption{
    \textbf{Transient $\ell_1$ perturbations reveal a similar reorganization of the generalization-time response.}
    A representative run from the $\ell_1$-regularized grokking setting of \citet{junior2025grokking} is shown.
    \textbf{(a)} Baseline training and test accuracy.
    \textbf{(b)} Generalization-time shifts under transient changes in $\ell_1$ regularization strength.
    \textbf{(c,d)} Test- and training-cross-entropy barriers between perturbed and baseline branches at matched generalization checkpoints. The test-loss barrier remains close to zero throughout most of the scan, whereas the training-loss barrier retains a clear late-stage collapse similar to that observed in the WD experiments.
    }
    \label{fig:l1_response}
\end{figure}
The $\ell_1$ response shows a qualitatively similar reorganization. A directional response to regularization pressure emerges before visible generalization: increasing $\ell_1$ regularization accelerates generalization, whereas decreasing it delays generalization. However, a clear dose-dependent ordering of the response appears only later, closer to the onset of generalization. The training-loss barrier retains a qualitatively similar late-stage collapse to that observed in the WD experiments, whereas the test-loss barrier remains close to zero throughout most of the scanned interval.

\section{Discussion}

Gradient-based optimization can favor particular solutions even when many solutions fit the training data equally well, reflecting its implicit bias \citep{soudry2018implicit,gunasekar2018characterizing}. In grokking, this preference can continue to change over training, for example from a kernel predictor to a minimum-norm solution \citep{lyu2024dichotomy}. Our results suggest that, in three grokking tasks, long before visible generalization, local perturbations become less able to redirect solution selection, while retaining a systematic influence on generalization timing.
\begin{figure}[!htbp]
    \centering
    \includegraphics[width=0.3\columnwidth]{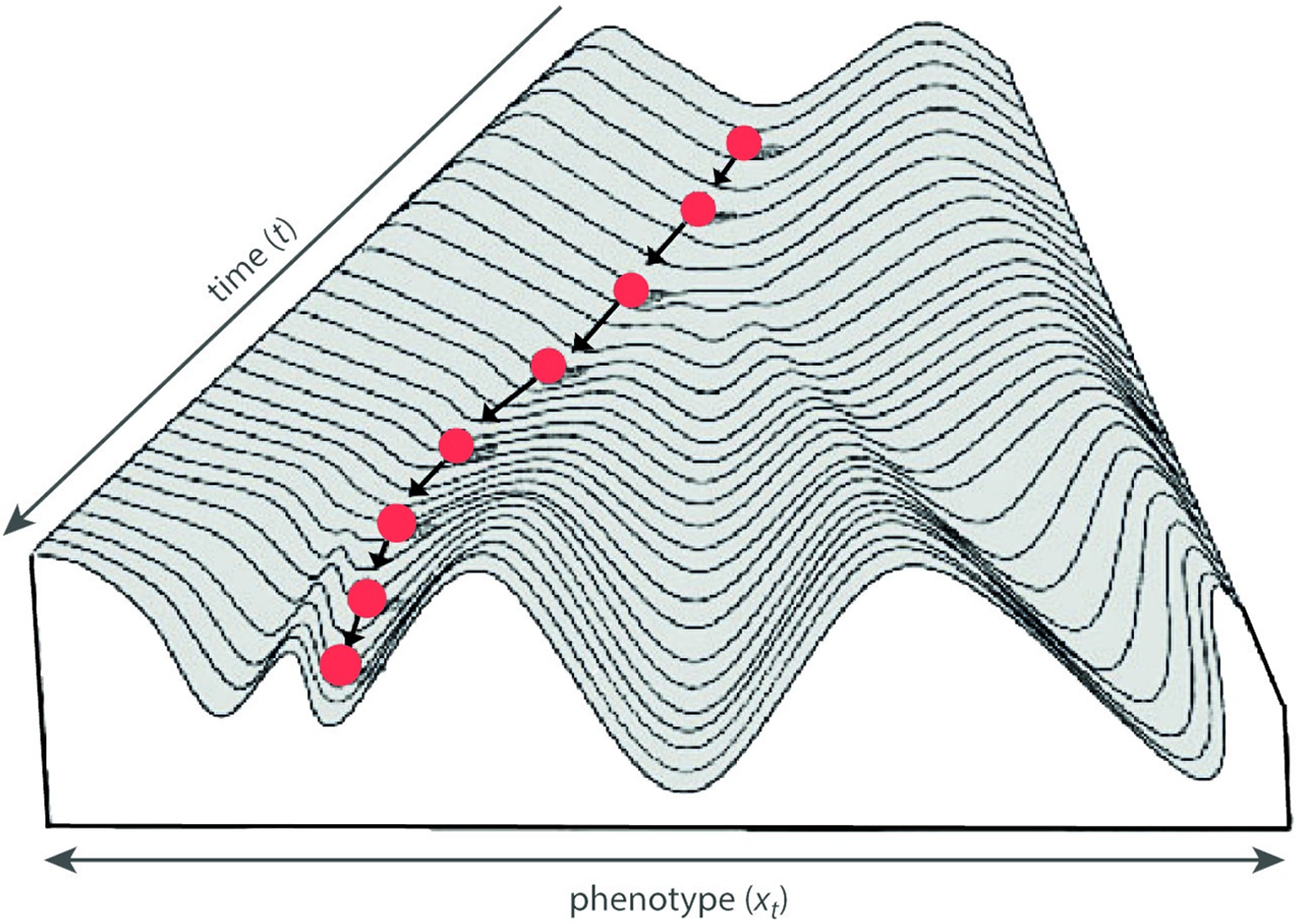}
   \caption{\textbf{Waddington's epigenetic landscape. Reproduced from Fig.~1 of \citet{mitteroecker2021canalization} under CC BY 4.0.}
Developing cells are depicted as balls rolling down valleys that become deeper and narrower over time, making perturbations increasingly less able to change their eventual direction. This landscape provides an intuitive analogy for the canalization of solution selection described here.}
    \label{fig:waddington_landscape}
\end{figure}
As shown in Figure~\ref{fig:waddington_landscape}, Conrad Waddington, who introduced the concept of epigenetics, used the epigenetic landscape to describe cell development. In 1942, Waddington introduced the term \textit{canalization} to describe how developmental outcomes become increasingly robust to perturbations as development proceeds; the stability of the developmental trajectory itself was later described as \textit{homeorhesis} \citep{waddington1942canalization,waddington1957strategy}. In this landscape, a developing cell is represented as a ball rolling down a valley. As the valley becomes deeper and narrower, perturbations become increasingly less able to change the ball's lateral direction. This provides an intuitive analogy for our results. Early perturbations can still change where the ball goes, whereas later perturbations have less effect on its lateral direction but can still speed up or slow down its motion along the valley, causing it to reach the endpoint earlier or later.

The ordered shifts in generalization timing can be viewed in the context of how regularization continues to shape solution selection after training fit. Recent theoretical work has provided a clear picture of the separation of timescales in grokking. Boursier et al. describe an initial phase in which the model rapidly fits the training data, followed by a slower phase in which small weight decay drives motion along a manifold of critical points of the training loss while gradually reducing the $\ell_2$ norm of the parameters \citep{boursier2025grokking}. In overparameterized ridge regression, \citet{xu2026togrok} provide an end-to-end proof of all three stages of this behavior: the model fits the training data early, poor generalization persists long after, and the model eventually generalizes. This picture extends beyond $\ell_2$ norm reduction. \citet{junior2025grokking} show that regularization favoring sparsity or low rank can also induce grokking when these properties support generalization on the task. Together, these studies suggest that regularization guides the direction of solution selection after training fit, while the match between its preferred properties and the task determines whether this evolution promotes generalization.

Our results are consistent with this picture of regularization continuing to shape solution selection after training fit. As WD pulses are applied later in training, the test-loss barriers between the perturbed and baseline generalization checkpoints decrease toward zero. Stable, dose-ordered shifts in generalization time persist even when these test-loss barriers are close to zero.  This pattern resembles slow drift within a connected solution region, where transient changes in regularization strength can even modulate generalization time approximately linearly near the end of the pre-generalization plateau. 

\begin{figure*}[t]
    \centering
    \includegraphics[width=0.6\columnwidth]{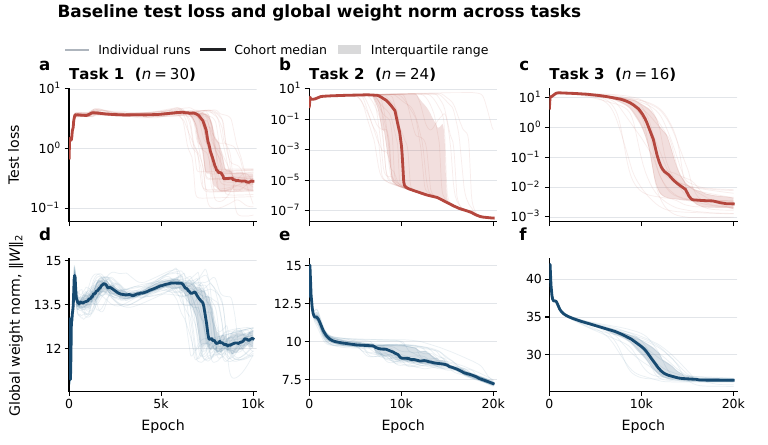}
    \caption{
    \textbf{Baseline test loss (top) and global weight norm (bottom) across Tasks\~1--3.}
   Task~1 shows little sustained norm decrease until test loss begins to improve, whereas Tasks~2 and~3 exhibit clearer norm contraction during late training. All quantities are evaluated every 25 epochs.
    }
    \label{fig:task1_baseline_dynamics}
\end{figure*}

As \citet{junior2025grokking} show, regularization can induce grokking by favoring properties suited to the task, such as sparsity or low rank. Under such regularization, generalization can improve even without a decrease in the overall Euclidean norm. In modular addition (our Task~3), the weight norm falls before generalization becomes visible \citep{liu2023omnigrok}, while in sparse parity (our Task~2), most neuron norms decay as a small generalizing subnetwork is amplified \citep{merrill2023tale}. In our parity-match task, however, the baseline weight norm changes very little over most of the plateau and begins a sustained decline only around the time the test loss starts to improve (Figure~\ref{fig:task1_baseline_dynamics}). The ordered shifts in generalization timing induced by transient changes in WD appear well before this decline. These observations show that regularization can already exert an ordered influence on generalization timing while the baseline weight norm remains nearly constant.
During this plateau in Task~1, AdamW weight decay continuously pulls the parameters toward zero, while the loss-driven updates roughly offset its effect on the global norm. The system enters the dose-ordered regime while this balance is still maintained. Later, the balance shifts, shrinkage begins to dominate, and the global norm decreases steadily, with visible generalization emerging around the onset of this decline. This pattern also helps explain why typical grokking trajectories are relatively rare in Task~1 (see Appendix~\ref{app:tasks_and_baselines}). A long pre-generalization plateau in this task requires the norm-decreasing and norm-increasing effects to remain balanced for a long time, with this balance shifting only later in training.

Our interventions help identify when training enters a regime in which regularization pressure toward simpler solutions suited to the task appears to become a major determinant of generalization timing. This regime is marked by stable, dose-ordered timing responses that persist even when the perturbed and baseline generalization solutions are linearly connected. Its onset and its relation to norm reduction and solution connectivity differ across tasks.
A checkpoint can be characterized by its current parameters, representations, or output function, but also by how controlled perturbations applied from that state affect subsequent training; the latter captures information about the local future dynamics \citep{golatkar2019time,frankle2020linear,altintas2025butterfly}. In our Task~1 setting, standard progress measures---including restricted loss \citep{nanda2023progress}, parameter movement, and feature amplification \citep{barak2022hidden}---show little signal during the plateau well before generalization, whereas the WD perturbation response already develops a stable structure before visible generalization (see Appendix~\ref{app:progress_measures}). In Task~1, perturbation responses therefore reveal a change in training dynamics well before it becomes visible in these standard progress measures.

Our experiments are limited to small algorithmic tasks, so it remains unclear whether similar changes in perturbation response can be observed in broader optimization settings. In more complex tasks, grokking curves may rise in multiple stages or show substantial fluctuations, making it difficult to define a single generalization time. The long and clearly defined pre-generalization plateau required by our intervention may also be absent. The three tasks studied here all exhibit a single, relatively sharp generalization transition, providing a clear temporal reference for the pulse phase, the onset of ordering, and the collapse of the barrier. Extending this intervention framework to more complex tasks will therefore require a more general way to define the generalization transition.

We use exactly the same pulse parameters across all three WD perturbation tasks, including the same window length $\tau=500$, ten perturbation doses, and scan stride, without tuning the intervention for individual tasks. Although the baseline weight decay ranges from $1.5$ to $5.0$ and the model architecture ranges from an MLP to an embedding model with a factorized head, the same intervention produces broadly similar response structures across the three tasks. This suggests that, within the settings studied here, the observed dynamical structure does not depend on task-specific tuning of the intervention.

Overall, we combine the temporal separation provided by grokking with transient changes in regularization strength to study how the effects on solution selection and generalization timing evolve over training. In the grokking tasks studied here, these perturbations become increasingly less able to steer training toward a linearly separated generalization checkpoint, while their effects on generalization time remain ordered and adjustable. We refer to this pre-generalization reorganization as the \textit{canalization} of grokking solution selection.

\section*{LLM Usage Claim}
We used LLMs for language polishing, coding and documentation assistance, and a reproducibility sanity check—where an AI agent tried to replicate our experiments from the paper description alone to see if our methods were written clearly enough. The agent's output was checked by us; it did not replace our own verification. All ideas, experiments, and conclusions are ours. 

\section*{Code availability}
Code, numerical results, and figure source data are provided in the supplementary material. Model checkpoints are available from the authors upon request.

\bibliographystyle{unsrtnat}
\begingroup
\small
\microtypesetup{protrusion=false}
\bibliography{references}
\endgroup

\clearpage
\section*{Appendix}
\appendix
\section{Tasks and Baseline Training}
\label{app:tasks_and_baselines}
\subsection{Three Tasks}

We use three algorithmic tasks that exhibit delayed generalization. All models are trained with full-batch updates, with one parameter update per epoch and no learning-rate scheduler. AdamW is used for all tasks; except for Task~3, the momentum parameters are set to $(\beta_1,\beta_2)=(0.9,0.999)$. Training and test metrics are evaluated every $25$ epochs until test accuracy reaches $0.60$, and every epoch thereafter.

\begin{center}
\captionof{table}{\textbf{Scale of the WD-pulse intervention analysis across tasks.}
Each pulse start is evaluated with 10 WD perturbation magnitudes, giving 10 intervention branches per pulse start.}
\label{tab:intervention_scale}

\begin{tabular}{lrrrr}
\toprule
Task & Runs & Pulse starts & Avg.\ per run & Branches \\
\midrule
Task~1 (parity match)              & 30 & 8,119  & 271 & 81,190  \\
Task~2 (sparse parity)             & 24 & 8,975  & 374 & 89,750  \\
Task~3 (factored modular addition) & 16 & 6,832  & 427 & 68,320  \\
\midrule
Total                              & 70 & 23,926 & --- & 239,260 \\
\bottomrule
\end{tabular}
\end{center}
\paragraph{Task~1: Parity match.}
Each example consists of two integer indices $(i,j)$, where $i,j\in\{0,\ldots,47\}$. The two indices are separately encoded as 48-dimensional one-hot vectors and concatenated into a 96-dimensional input. The label indicates whether the two indices have the same parity:
\[
y=\mathbf{1}\!\left[(i\bmod 2)=(j\bmod 2)\right].
\]

The full input space contains $48\times48=2304$ ordered pairs. Using the fixed data seed \texttt{data\_seed=21}, we first randomly select $75\%$ of the inputs as the candidate training pool, and then draw 304 samples with replacement from this pool to construct the training set. Because sampling is performed with replacement, these 304 training entries correspond to 267 distinct input pairs; repeated pairs contribute to the full-batch loss according to their multiplicity. The test set contains 1024 samples drawn from input pairs that do not appear in the training set, with balanced binary labels.

The model is a 4-hidden-layer MLP with 128 units per layer and $\tanh$ activations. The output layer produces a single binary-classification logit. The training loss is binary cross-entropy with logits. The learning rate is $10^{-3}$, the baseline weight decay is $5.0$, and the model is trained for $10{,}000$ epochs.

\paragraph{Task~2: Sparse parity.}
This task is based on the sparse-parity grokking setup of Merrill et al.~\citep{merrill2023tale}. Each example begins with a binary vector
\[
b=(b_0,\ldots,b_{49})\in\{0,1\}^{50},
\]
which is then transformed into $-1/+1$ features by
\[
\tilde b=2b-1
\]
before being passed to the model. The label depends only on three fixed relevant coordinates:
\[
y=
\left(
\sum_{k\in\mathcal R} b_k
\right)\bmod 2,
\qquad
\mathcal R=\{10,26,44\}.
\]
Indices are zero-based, corresponding to the 11th, 27th, and 45th coordinates of the input vector. The remaining 47 coordinates are distractor features that do not contribute to the label. The relevant coordinates are selected using the fixed \texttt{relevant\_bit\_seed=21}, and the training and test sets are generated using the fixed \texttt{data\_seed=21}.
The training and test sets contain 550 and 1000 non-overlapping input vectors, respectively, with balanced binary labels in the training set. The model is a 4-hidden-layer MLP with 128 units per layer and $\tanh$ activations. The output layer produces a single binary-classification logit. The training loss is binary cross-entropy with logits. The learning rate is $10^{-3}$, the baseline weight decay is $1.5$, and the model is trained for $20{,}000$ epochs.

\paragraph{Task~3: Factored modular addition.}
This task is based on the factored modular-addition setup of Google PAIR~\citep{pearce2023memorize}. Each input is a pair $(x,y)$ of integers modulo 67, with target class
\[
z=(x+y)\bmod 67.
\]
Using the commutativity of addition, we retain only pairs satisfying $0\leq x\leq y\leq 66$, giving a full sample space of
\[
\frac{67\times68}{2}=2278
\]
examples. Using the fixed data seed \texttt{data\_seed=165}, we randomly shuffle the full sample space and use 570 examples for training, approximately $25\%$ of the full space, with the remaining 1708 examples used for testing.

The model uses a shared integer embedding matrix
\[
E\in\mathbb{R}^{67\times500}.
\]
Let $e_x,e_y\in\mathbb{R}^{500}$ denote the embeddings corresponding to integers $x$ and $y$. Both inputs are mapped into the hidden space through the shared projection matrix
\[
P_{\mathrm{in}}\in\mathbb{R}^{500\times128},
\]
summed, and passed through a ReLU activation:
\[
h=
\operatorname{ReLU}
\left(
e_xP_{\mathrm{in}}+e_yP_{\mathrm{in}}
\right).
\]
The hidden representation is then mapped back to the embedding space through
\[
P_{\mathrm{out}}\in\mathbb{R}^{128\times500},
\]
and tied unembedding is used to produce 67-dimensional output logits:
\[
\ell=hP_{\mathrm{out}}E^\top.
\]
Thus, the input integer representations and output class weights share the same embedding matrix $E$. No bias terms are used in the hidden or output layers.

The training objective is multiclass cross-entropy scaled by a factor of $1/67$. The model is trained with AdamW using a learning rate of $10^{-3}$, a baseline weight decay of $2.0$, and momentum parameters $(\beta_1,\beta_2)=(0.9,0.98)$, for a total of $14{,}000$ epochs.

\subsection{Baseline Run Selection}

The initial screening pool for Task~1 contains 1,000 baseline runs. Let $T_{q,d}^{(3)}$ denote the first time the accuracy on split $d\in\{\mathrm{train},\mathrm{test}\}$ reaches $q$, with the following three evaluation points also remaining at or above $q$. A run is included in the main analysis only if
\[
\begin{aligned}
T_{0.60,\mathrm{test}}^{(3)}
-
T_{0.99,\mathrm{train}}^{(3)}
&\geq 5000,\\
T_{0.75,\mathrm{test}}^{(3)}
-
T_{0.55,\mathrm{test}}^{(3)}
&< 1000
\end{aligned}
\]
and its test accuracy at the end of training is at least $0.95$. These three criteria require, respectively, a sufficiently long pre-generalization plateau, a concentrated and clearly defined generalization transition, and high test accuracy within the training limit.

Among the 1,000 runs, 177 reach a final test accuracy of at least $0.95$, and 105 satisfy all criteria. Because a full intervention scan requires thousands of perturbed branches per run, we use the first 30 qualifying runs in increasing seed order for the main analysis. The intervention scale across all three tasks is summarized in Table~\ref{tab:intervention_scale}. Figure~\ref{fig:wd_response_across_runs_tasks}(a) shows the first three runs satisfying the criteria (seeds 2, 4, and 9). The example runs for Task~2 and Task~3 shown in Figure~\ref{fig:wd_response_across_runs_tasks}(c,d) and Figure~\ref{fig:mode_connectivity_response} use seed 9 and seed 1, respectively.
For Task~2, baseline screening adds a monotonicity requirement to the three Task~1 criteria. Starting from the first time test accuracy reaches $0.60$, every subsequent evaluation point must have accuracy no lower than the previous point, excluding trajectories with reversals or multi-stage fluctuations during generalization (until test accuracy $\geq 0.99$). The screening pool contains 508 baseline runs (seeds 1--508), of which 24 satisfy all criteria and are included in subsequent analyses.

Task~3 consistently exhibits clear grokking across different initializations, so no additional screening is applied. We directly use seeds 1--16 for the cross-initialization analysis.

\begin{figure}[t]
    \centering
    \includegraphics[width=\linewidth]{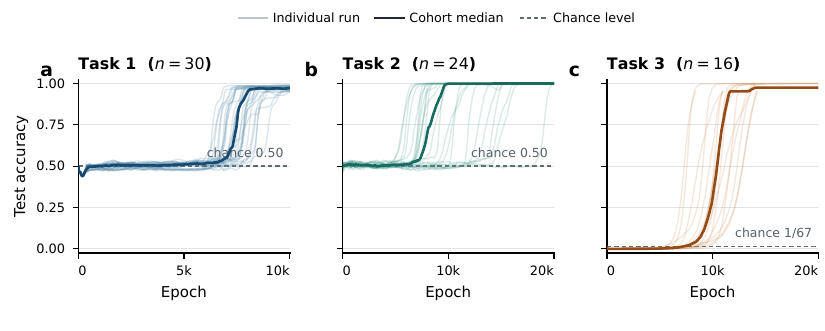}
    \caption{\textbf{All baseline generalization trajectories included in the main analysis.}
    We show 30 runs for Task~1, 24 runs for Task~2, and 16 runs for Task~3. Thin lines show baseline test accuracy for individual initializations, thick lines show the median across runs, and dashed lines indicate chance level.}
    \label{fig:baseline_runs}
\end{figure}

Figure~\ref{fig:baseline_runs} summarizes all baseline generalization trajectories included in the main analysis.

\section{Dose--Response Linearity and Weight-Norm Analysis}
\label{app:dose_response_analysis}

\subsection{Dose--Response Linearity Analysis}
\label{app:dose_response_linearity}

To quantify when the WD response changes from an unordered pattern to a stable dose relationship, we examine whether the generalization-time shift $\Delta T_{0.90}$ varies approximately linearly with WD perturbation magnitude at each pulse start. Because the generalization time scale varies substantially across initializations, we first align the pre-generalization plateau of each run using a common training-phase coordinate.

Let $t_{\mathrm{fit}}$ denote the first checkpoint at which training accuracy reaches $1.0$, and let $T_{0.60}^{0}$ denote the first time the baseline test accuracy reaches $0.60$. The phase of pulse start $t_0$ is defined as
\[
\phi(t_0)
=
\frac{t_0-t_{\mathrm{fit}}}
{T_{0.60}^{0}-t_{\mathrm{fit}}}.
\]
Thus, $\phi=0$ corresponds to the first time training accuracy reaches $1.0$, and $\phi=1$ corresponds to the first time baseline test accuracy reaches $0.60$. We divide the phase coordinate into bins of width $0.01$ so that runs can be compared at similar stages of training.

Positive and negative WD perturbations are analyzed separately at each pulse start. Each perturbation direction includes five magnitudes,
\[
|\Delta\lambda|
\in
\{0.05,0.10,0.15,0.20,0.25\}.
\]
We fit
\[
\Delta T_{0.90}
=
\beta_0+\beta_1|\Delta\lambda|+\varepsilon,
\]
and use the squared Pearson correlation coefficient, $R^2$, to measure how closely the five responses follow a linear dose--response relationship. Because each fit contains only five dose points, random assignments of responses to doses can produce nonzero $R^2$ values even when no true dose relationship exists. For five fixed doses, averaging $R^2$ over all permutations of the response values gives
\[
\overline{R^2}_{\mathrm{perm}}=\frac{1}{4}.
\]
We therefore define the chance-corrected linearity as
\[
R^2_{\mathrm{corr}}
=
\frac{R^2-1/4}{1-1/4}.
\]
On this scale, $R^2_{\mathrm{corr}}=0$ corresponds to the mean value under random permutations, while $R^2_{\mathrm{corr}}=1$ corresponds to a perfectly linear dose--response.

We compute $R^2_{\mathrm{corr}}$ for a pulse start only when all five doses in the corresponding perturbation direction yield valid $\Delta T_{0.90}$ values and the responses have nonzero variance. If any branch fails to reach $0.90$ test accuracy within the continuation limit, that pulse start is treated as missing for the corresponding perturbation direction.

To summarize results across runs, each pulse start is first assigned to its corresponding phase bin. If multiple pulse starts from the same run fall into the same bin, we take the median $R^2_{\mathrm{corr}}$ within that bin without temporal interpolation, and then average across runs. For each phase bin, different random initializations are treated as independent samples, and valid runs are resampled with replacement $4{,}000$ times. For each resampled set of runs, we recompute the mean $R^2_{\mathrm{corr}}$ across runs. The $2.5\%$ and $97.5\%$ percentiles of these means are reported as the $95\%$ confidence interval. The final four phase bins ($\phi\geq0.965$) contain fewer than 15 valid runs and are therefore not shown; the plotted curves end at $\phi=0.955$.

\subsection{Baseline Weight Norm}
\label{app:baseline_weight_norm}

To compare the evolution of the WD dose--response with changes in parameter scale, we compute the weight norm from baseline checkpoints. Let $\{W_\ell\}$ denote all non-bias weight tensors in the model. The overall weight norm is defined as
\[
\lVert W\rVert_2
=
\left(
\sum_\ell \lVert W_\ell\rVert_F^2
\right)^{1/2}.
\]
For each run, we use the weight norm at the first checkpoint at which training accuracy reaches $1.0$ as the reference value, and divide the weight norm at each checkpoint by this reference to obtain the relative weight norm.

We aggregate the weight-norm measurements across runs using the same procedure as in the dose--response linearity analysis. We use the same phase coordinate and phase bins of width $0.01$. If multiple checkpoints from the same run fall into the same bin, we first take their median and then average across runs. For each phase bin, different random initializations are treated as independent samples, and valid runs are resampled with replacement $4{,}000$ times. For each resampled set of runs, we recompute the mean relative weight norm across runs. The $2.5\%$ and $97.5\%$ percentiles of these means are reported as the $95\%$ confidence interval. As in the dose--response linearity analysis, the final four phase bins ($\phi\geq0.965$) contain fewer than 15 valid runs and are not shown, so the plotted curve ends at $\phi=0.955$.

\subsection{Linearity--Weight-Norm Synchrony}
\label{app:linearity_norm_synchrony}

To test whether the emergence of dose--response linearity is synchronized with changes in the baseline weight norm, we compare how the two quantities change across training phase within each run. The synchrony analysis is restricted to $\phi\in[0,1]$, from the first time training accuracy reaches $1.0$ to the first time baseline test accuracy reaches $0.60$.

For each phase bin, we average the $R^2_{\mathrm{corr}}$ values for positive and negative WD perturbations and denote the result by $\bar R^2_{\mathrm{corr}}$; this quantity is defined only when both perturbation directions are valid. Let $r$ denote the baseline relative weight norm in the same bin. For each pair of consecutive valid phase bins, we compute the changes in $\bar R^2_{\mathrm{corr}}$ and $\ln r$, and then calculate the Spearman correlation between these two sets of changes within each run.

All 30 runs contain sufficient valid consecutive-bin differences and are therefore included in the synchrony analysis, with 24--99 difference pairs per run. We summarize the resulting run-level Spearman correlations by their median. Treating different random initializations as independent samples, we resample the 30 runs with replacement $4{,}000$ times. For each resampled set of runs, we recompute the median Spearman correlation across runs and report the $2.5\%$ and $97.5\%$ percentiles of these medians as the $95\%$ confidence interval.

\section{WD Pulse Implementation and Test-Loss Barriers}
\label{app:wd_implementation}

\subsection{Implementation of WD Pulse Interventions}

We implement the sliding-window protocol defined in Section~\ref{sec:dynamical_probe} as follows. For each pulse start $t_0$, we resume training from the complete training state $\mathcal{S}_{t_0}$ saved at that time. This state includes all model parameters, the AdamW optimizer state (first- and second-moment estimates), the random-number-generator state, and the data-sampling state. All perturbed branches at the same $t_0$ start from exactly the same state and differ only in the weight-decay coefficient. During the pulse window ($t_0<t\leq t_0+\tau$), the AdamW weight decay is set to $\lambda_0+\Delta\lambda$; after the window ends, it is restored to $\lambda_0$. The learning rate, momentum parameters, and batch configuration remain unchanged throughout. In AdamW, weight decay enters the parameter update through the decoupled term $-\eta\lambda\theta$, so the WD pulse directly changes this parameter-shrinkage term without changing the loss function or gradient computation.

Pulse starts $t_0$ begin at the checkpoint where training accuracy first reaches $1.0$ and are scanned along the plateau every $25$ epochs. The upper scan limit is $T^{0}_{0.95}-1{,}000$, where $T^{0}_{0.95}$ is the first time baseline test accuracy reaches $0.95$. The pulse duration is fixed at $\tau=500$ epochs, with perturbation magnitudes
\[
\Delta\lambda\in\{\pm0.05,\pm0.10,\pm0.15,\pm0.20,\pm0.25\}.
\]
The baseline weight-decay values are $5.0$ for Task~1, $1.5$ for Task~2, and $2.0$ for Task~3 (see Appendix~\ref{app:tasks_and_baselines}).

After the pulse ends, each perturbed branch returns to the baseline training conditions and continues training until epoch $40{,}000$. We evaluate test accuracy every $25$ epochs while it is below $0.60$, and every epoch after it reaches $0.60$ to increase temporal resolution around the generalization transition. $T_\alpha(t_0,\Delta\lambda)$ is defined as the first evaluation time at which the perturbed branch reaches the test-accuracy threshold $\alpha$. If a branch does not reach the threshold before the training limit, it is marked as not reaching the threshold. The corresponding response and barrier are not computed, and the location is shown in gray in the figures.

\subsection{Test-Loss Barrier Computation}

For each perturbed branch, we compare two corresponding generalization checkpoints. Endpoint $A$ is the baseline parameter checkpoint at the first time test accuracy reaches $95\%$, and endpoint $B$ is the perturbed branch checkpoint at the first time it reaches the same accuracy. All perturbed branches within a run are compared against the same baseline endpoint $A$. Following the linear mode connectivity setup of \citet{frankle2020linear}, we linearly interpolate between the two endpoints in parameter space:
\[
\theta(s)=(1-s)\,\theta_A+s\,\theta_B,
\]
using the interpolation grid
\[
\mathcal G=\left\{0,\frac{1}{50},\ldots,\frac{49}{50},1\right\},
\]
which contains $51$ equally spaced points including the endpoints. For each interpolated model, we compute the test loss $L_{\mathrm{path}}(s)$ on the full test set.

Following \citet{entezari2022role}, we define the test-loss barrier as
\[
B(A,B)
=
\max_{s\in\mathcal G}
\Big[
L_{\mathrm{path}}(s)
-
\big((1-s)L_A+sL_B\big)
\Big].
\]
This quantity is the maximum increase in test loss along the interpolation path relative to the linear interpolation of the two endpoint losses. Unlike the endpoint-mean baseline used by \citet{frankle2020linear}, this definition does not count the linear loss difference between unequal endpoints as part of the barrier.

The two endpoints share the same training trajectory before the WD pulse, so we directly interpolate their parameters without permutation alignment. None of the models contain BatchNorm or other components that depend on running statistics, so the interpolated models can be evaluated directly. If a perturbed branch does not reach $95\%$ test accuracy within the observation limit, the corresponding generalization checkpoint does not exist. That $(t_0,\Delta\lambda)$ location is treated as missing and shown in gray in the figures.

\section{\texorpdfstring{$\ell_1$}{l1}-Regularized Grokking}
\label{app:l1}

We use the modulo-97 addition setting and matrix-product architecture in the released implementation of \citet{junior2025grokking}, with 40\% of the input table used for training. Training uses Adam with a fixed learning rate of $10^{-3}$, minibatches of size $2{,}048$, and no optimizer weight decay. The baseline objective is cross-entropy with an $\ell_1$ penalty on the model weights, using $\lambda_0=5\times10^{-6}$.
For the transient intervention, we replace $\lambda_0$ for 500 updates by $\lambda_0(1+r)$, with
\[
r\in\{\pm0.15,\pm0.30,\pm0.45,\pm0.60,\pm0.75\},
\]
and then restore the baseline coefficient. Pulse starts are scanned every 50 updates up to $T^0_{0.95}-500$. Timing responses are evaluated at $\alpha=0.95$. The continuation and endpoint-matching procedures otherwise follow Appendix~\ref{app:wd_implementation}. Train- and test-loss barriers are evaluated separately using cross-entropy without the $\ell_1$ penalty.

\section{Update-Direction Analysis}
\label{app:update_directions}

This appendix complements the test-loss barrier analysis in Section~\ref{sec:barrier_collapse} by examining the evolution of parameter-update directions after WD pulses. We compare early and ordering-onset pulses in three Task~1 runs (seeds 2, 4, and 9, the same runs shown in Figure~\ref{fig:wd_response_across_runs_tasks}(a)), following each perturbed trajectory from the end of its pulse.

\paragraph{Ordering onset and pulse selection.}
For every candidate pulse start, we examine four test-accuracy thresholds,
$\alpha\in\{0.60,0.75,0.90,0.95\}$,
for both positive and negative WD perturbations. This gives eight dose--response sequences, each containing five perturbation magnitudes. A checkpoint is classified as ordered only if all eight sequences have complete responses, the adjacent nonzero response changes have a consistent direction, and the absolute Spearman correlation between dose and $\Delta T_\alpha$ is at least $0.8$ for all eight sequences. The first checkpoint along the scan axis that satisfies these conditions is defined as the ordering onset.

For each run, we select a pulse start at the ordering onset and an early pulse start $500$ epochs before it. Each pulse lasts $\tau=500$ epochs and uses the same ten WD doses as in Appendix~\ref{app:wd_implementation}. The comparison therefore includes the baseline and $20$ perturbed trajectories per run.

\paragraph{Update directions and observation window.}
For each trajectory, we estimate the update direction in the full parameter space at epoch $t$ using a centered difference and normalize it:
\begin{equation}
    u_t=
    \frac{\theta_{t+25}-\theta_{t-25}}
    {\|\theta_{t+25}-\theta_{t-25}\|_2}.
\end{equation}
Thus, $u_t$ describes the direction of parameter movement over a $50$-epoch interval, without retaining its magnitude. We sample these directions every $50$ epochs, beginning at the end of each pulse and continuing up to epoch $10{,}000$. Each panel reports the number of training epochs shown after the pulse ends. Markers identify the pulse end and $500$, $1{,}000$, and $2{,}000$ epochs after it, making elapsed training time explicit along the projected trajectories.

\paragraph{Joint PCA.}
For each run, we jointly fit a single PCA basis to the sampled direction vectors from the baseline and all early and ordering-onset branches. The vectors are jointly centered, and the first two principal components are used for visualization. Both panels within a run share the same PCA basis, coordinate range, and axis aspect ratio. The explained-variance ratio of each displayed component is reported on the corresponding axis.

\begin{figure}[t]
    \centering
    \includegraphics[width=\textwidth]{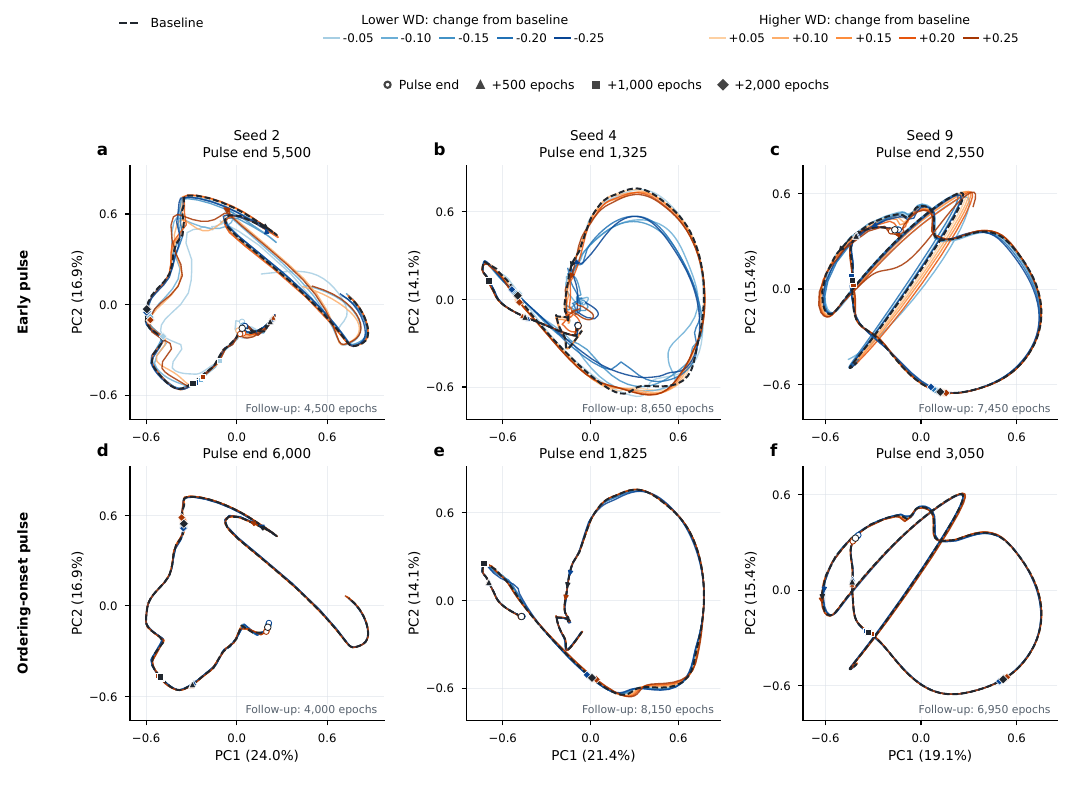}
    \caption{\textbf{Projected update-direction trajectories after early and ordering-onset WD pulses.}
Columns correspond to Task~1 seeds 2, 4, and 9; rows show early and ordering-onset pulses. Early pulses begin $500$ epochs before the corresponding ordering-onset pulses. Black dashed curves show the baseline; blue and orange curves show branches with decreased and increased WD, respectively. Trajectories are displayed from the pulse end, with circles, triangles, squares, and diamonds marking the pulse end and $500$, $1{,}000$, and $2{,}000$ epochs afterward. Each panel reports the number of training epochs shown after the pulse ends. Both panels within a run share a jointly fitted PCA basis, coordinate range, and axis aspect ratio; axis labels report the explained-variance ratios. Early branches show clearer deviations from the baseline, while ordering-onset branches follow largely overlapping paths in the displayed projection.}
    \label{fig:update_direction_pca}
\end{figure}

\paragraph{Results.}
Across the three runs, early pulse branches show clear deviations from the baseline in the PCA projection, whereas ordering-onset branches follow largely overlapping projected paths (Figure~\ref{fig:update_direction_pca}). This overlap is observed over several thousand epochs after the pulse ends. The early and ordering-onset pulse starts differ by only $500$ epochs, and both groups are tracked for comparable numbers of epochs after the pulse ends.

\FloatBarrier
\section{Comparison with Existing Progress Measures}
\label{app:progress_measures}

\begin{figure*}[t]
    \centering
    \includegraphics[width=\textwidth]{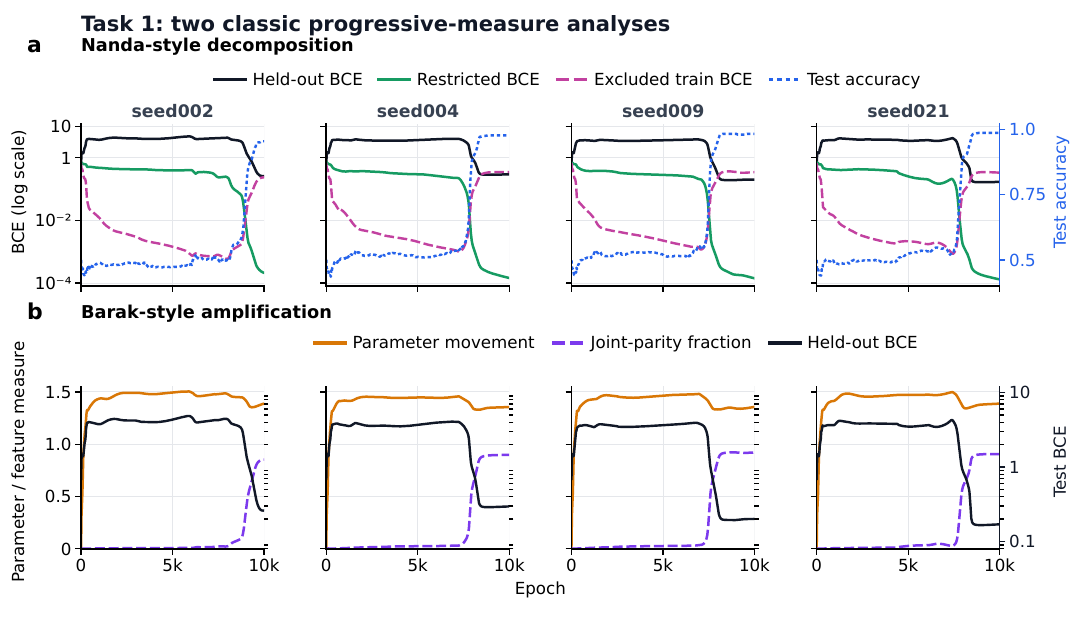}
\caption{\textbf{Existing grokking progress measures provide little advance signal during the Task~1 plateau.}
Each column corresponds to one clear-grokking run (seeds 2, 4, 9, and 21).
\textbf{(a)} Nanda-style restricted/excluded decomposition. Restricted test BCE remains near chance level throughout the plateau and begins to decrease only about $50$ epochs before the held-out test BCE begins to decrease (black curve); excluded train BCE continues to decrease and rises again near the generalization transition. The blue curve shows test accuracy (right axis).
\textbf{(b)} Barak-style parameter and feature amplification. Parameter movement reaches a stable level early in training and changes little thereafter; the Fourier energy fraction of the joint-parity mode in the fourth hidden layer remains near zero during the plateau and rises near the generalization transition as held-out test BCE decreases.}
\label{fig:task1_progress_measures}
\end{figure*}
Progress measures in the grokking literature ask whether observable changes related to generalization appear inside the model before test performance improves. If such signals emerge early during the plateau, hidden progress toward generalization can be tracked through static measurements of checkpoints. We test whether two representative classes of progress measures provide such an advance signal in Task~1 using four clear-grokking runs (seeds 2, 4, 9, and 21). Both classes of measures are computed directly from checkpoints saved during training. Results are shown in Figure~\ref{fig:task1_progress_measures}.

\paragraph{Restricted/excluded decomposition.}
\citet{nanda2023progress} decompose model outputs into a component aligned with the correct rule and a remaining component. The restricted model retains only the rule-aligned component, so a decrease in its test loss indicates that this component has begun to support generalization. The excluded model removes the rule-aligned component from the original output, and its train loss is used to characterize the remaining memorization component. For Task~1, the correct rule corresponds to a fixed parity checkerboard pattern over the $48\times48$ input grid, allowing this decomposition to be performed directly in output-function space.

We first arrange the model logits over all $2304$ inputs $(i,j)$ into a matrix $L_t\in\mathbb{R}^{48\times48}$, where matrix entry $(i,j)$ corresponds to the output on input pair $(i,j)$. We then remove the mean logit shared across all inputs to obtain the centered matrix
\[
C_t=L_t-\bar L_t.
\]
This removes the global bias while preserving variation in the output across input combinations.

The correct rule in Task~1 depends only on whether the two indices have the same parity, producing a fixed checkerboard pattern over the full input grid. We refer to this pattern as the joint-parity direction, corresponding to the interaction between the parity of the two inputs, and write it as a matrix with unit Frobenius norm:
\[
R[i,j]=\frac{(-1)^i(-1)^j}{48}.
\]
The projection coefficient of the model output onto this rule direction is
\[
a_t=\langle C_t,R\rangle.
\]
The corresponding rule-aligned component of the model output is therefore
\[
P_t=a_tR.
\]

Restricted logits are defined as $\bar L_t+P_t$, retaining only the global bias of the original model and the joint-parity rule component, and restricted BCE is computed on the test set. Excluded logits are defined as $L_t-P_t$, removing the joint-parity rule component from the original output, and excluded BCE is computed on the training set. Restricted BCE therefore tracks when the rule component begins to support predictions on unseen samples, while excluded BCE tracks changes in the remaining training-fit component after the rule component has been removed.

\paragraph{Barak-style amplification.}
\citet{barak2022hidden} study whether parameters and features associated with the target rule are gradually amplified before generalization appears. If similar hidden progress occurs in Task~1, corresponding signals should accumulate during the plateau. We record two quantities.

First, we measure parameter movement relative to initialization:
\[
\frac{\|\theta_t-\theta_0\|_2}{\|\theta_0\|_2},
\]
which measures the overall displacement of the model parameters from initialization.

Second, we examine whether the complete joint-parity rule gradually appears in the hidden-layer representations. We perform a two-dimensional Fourier decomposition of hidden-layer activations over the full $48\times48$ input grid. The modes $(24,0)$ and $(0,24)$ correspond to the parity patterns of the two individual inputs, while $(24,24)$ corresponds to the joint-parity rule formed by their interaction. Figure~\ref{fig:task1_progress_measures} reports the Fourier energy fraction of the joint-parity mode in the fourth hidden layer, defined as the sum of squared amplitudes at the $(24,24)$ mode across all hidden units divided by the total Fourier energy after excluding the constant $(0,0)$ mode.

\paragraph{Results.}
In Task~1, these standard progress measures do not show a clear signal of generalization substantially before it becomes visible. Restricted test BCE remains near chance level throughout almost the entire plateau and begins to decrease only about $50$ epochs before the held-out test BCE begins to decrease. Excluded train BCE continues to decrease during training and rises again around generalization. The Barak-style measures show a similar pattern. Parameter movement saturates early in training and changes little thereafter, while the Fourier energy fraction of the joint-parity mode in the fourth hidden layer remains near zero throughout the plateau and rises only around generalization. Thus, in Task~1, these static progress measures do not reveal clear hidden progress substantially before visible generalization. In contrast, the dose-dependent structure of the WD response is already established roughly $4{,}000$ epochs before the baseline crosses $\alpha=0.90$ (see Section~\ref{sec:wd_response_ordering}).
\FloatBarrier
\section{Additional Runs Across Random Initializations}
\label{app:cross_seed_analysis}

We repeat the analysis from Section~\ref{sec:barrier_collapse} across 30 Task~1 runs, 24 Task~2 runs, and 16 Task~3 runs. Endpoint selection, the number of interpolation points, and the definition of the test-loss barrier are the same as in Figure~\ref{fig:mode_connectivity_response} (see Appendix~\ref{app:wd_implementation}).

Figures~\ref{fig:cross_seed_barrier_wdr_task1}--\ref{fig:cross_seed_barrier_wdr_task3} show that the relationship reported in the main text is broadly reproduced across random initializations. As the pulse start moves later in training, the test-loss barrier between the perturbed and baseline generalization checkpoints decreases, while the directional and dose-dependent structure of $\Delta T_{0.95}$ persists. Task~1 shows substantial variation in perturbation sensitivity across initializations, and some runs contain many perturbed branches that do not reach $0.95$ test accuracy within the observation limit.

The three tasks differ in the degree of trajectory heterogeneity across initializations. Task~1 and Task~2 runs are selected using predefined baseline-screening criteria, whereas Task~3 uses seeds 1--16 without any seed screening. Despite this, Task~3 shows particularly consistent structure across initializations: the changes in both the WD timing responses and the test-loss barriers are highly consistent across seeds. This indicates that, at least in Task~3, the observed phenomenon does not depend on seed screening based on baseline trajectories.

\begin{figure*}[t]
    \centering
    \includegraphics[width=\textwidth]{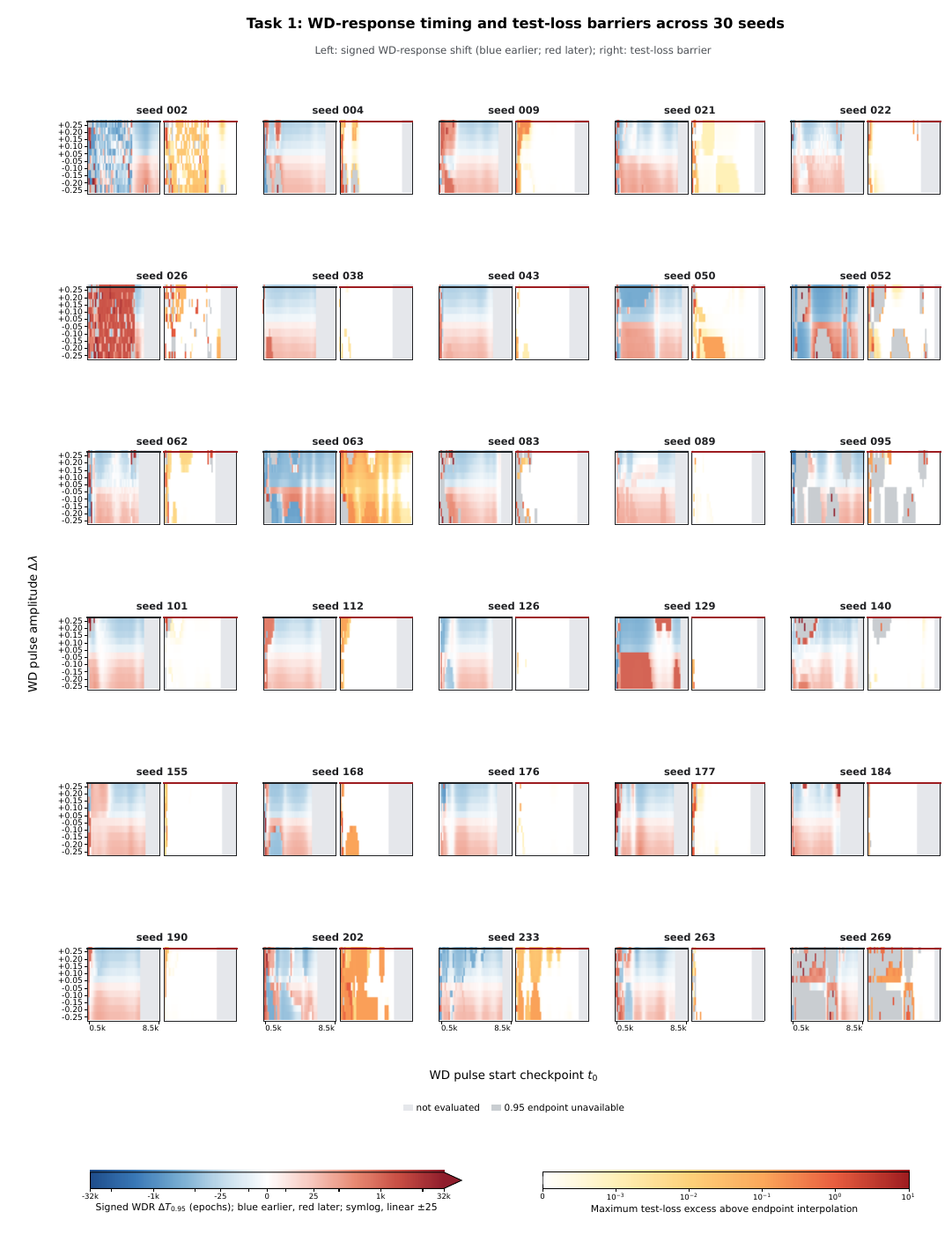}
    \caption{WD responses and test-loss barriers across 30 independent Task~1 initializations. Run-selection criteria are described in Appendix~\ref{app:tasks_and_baselines}.}
    \label{fig:cross_seed_barrier_wdr_task1}
\end{figure*}

\begin{figure*}[t]
    \centering
    \includegraphics[width=\textwidth]{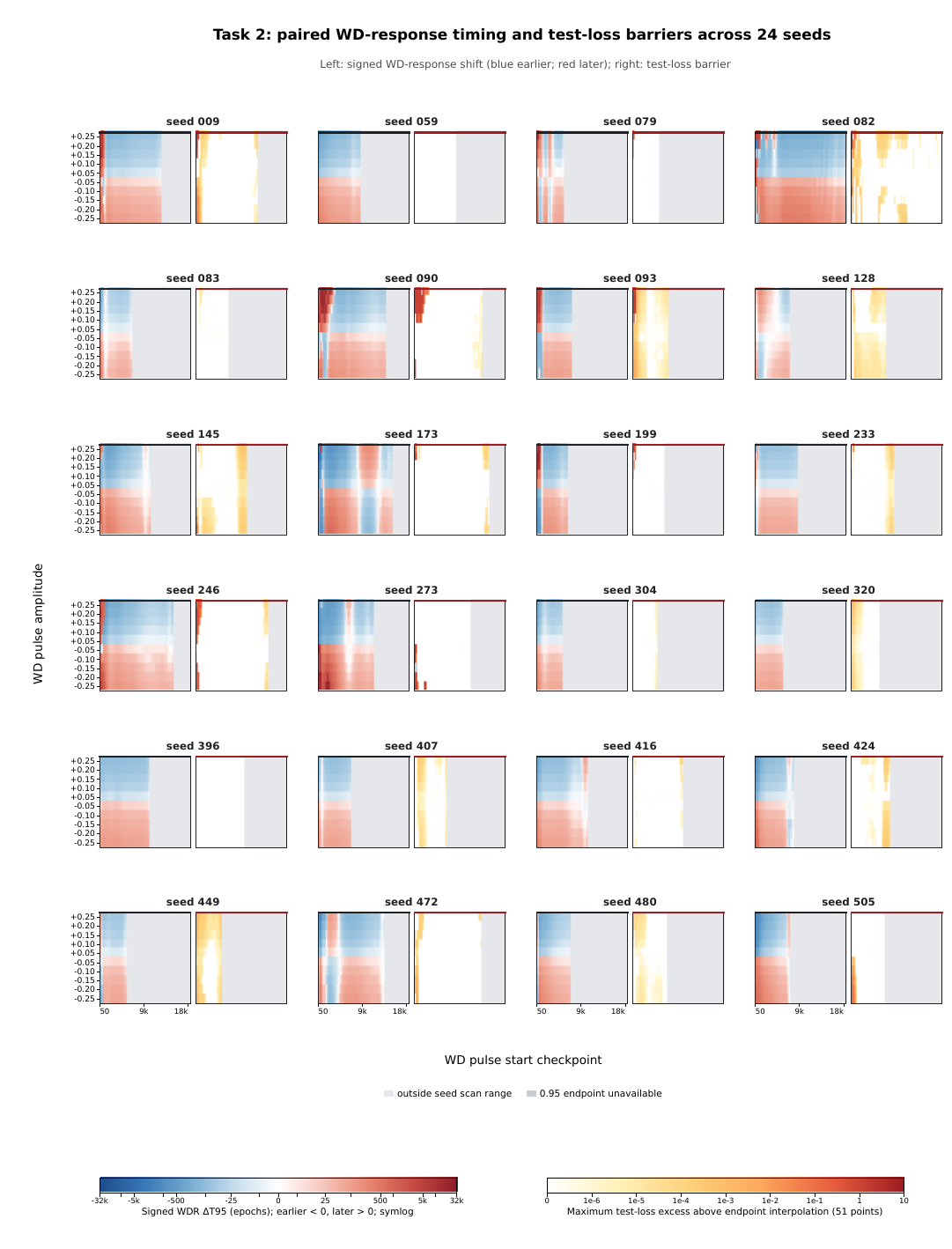}
    \caption{WD responses and test-loss barriers across 24 independent Task~2 initializations. Run-selection criteria are described in Appendix~\ref{app:tasks_and_baselines}.}
    \label{fig:cross_seed_barrier_wdr_task2}
\end{figure*}

\begin{figure*}[t]
    \centering
    \includegraphics[width=\textwidth]{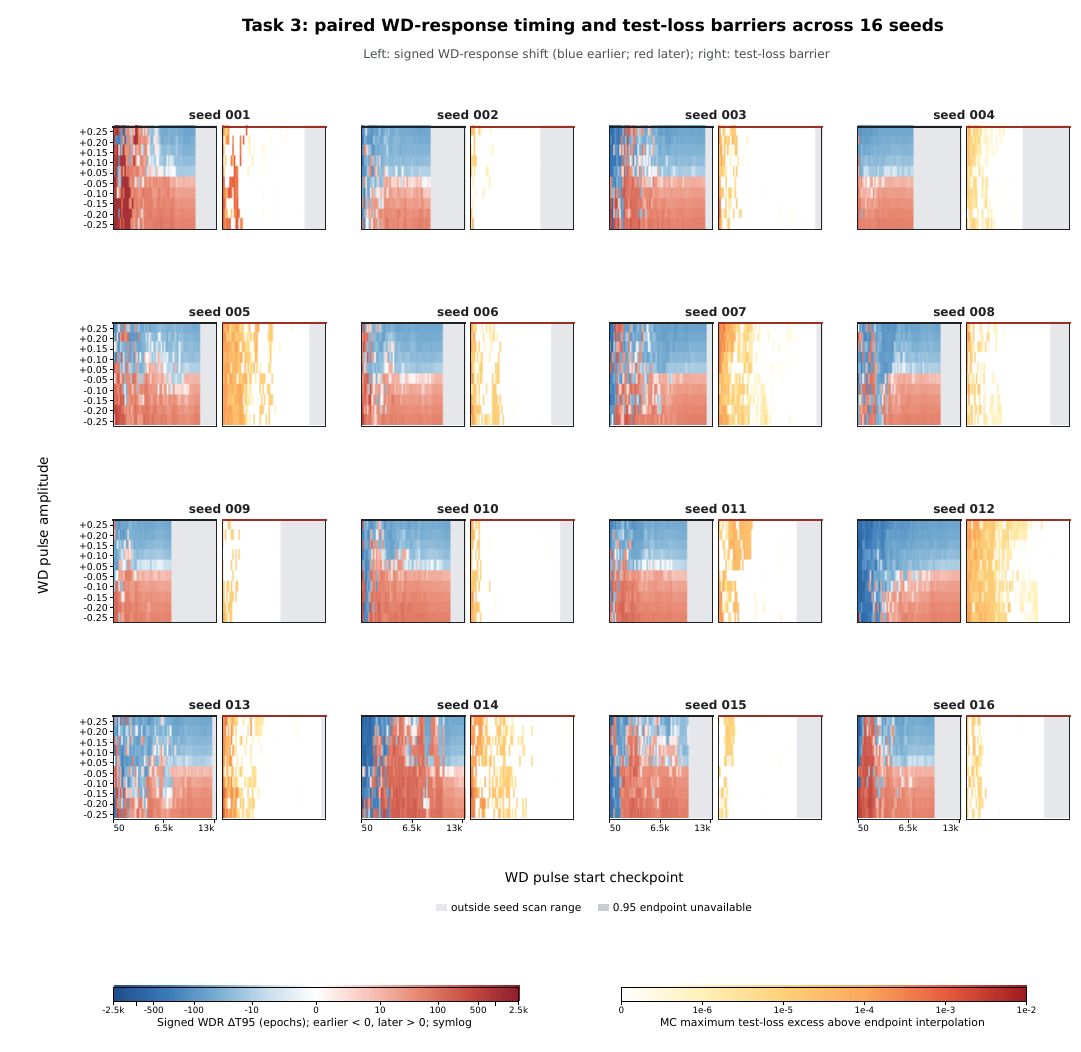}
    \caption{WD responses and test-loss barriers across 16 independent Task~3 initializations, corresponding to seeds 1--16.}
    \label{fig:cross_seed_barrier_wdr_task3}
\end{figure*}

\clearpage
\section{Additional Runs under \texorpdfstring{$\ell_1$}{l1} Regularization}
\label{app:l1_across_seeds}

\begin{figure*}[!htbp]
    \centering
    \includegraphics[width=0.9\textwidth]{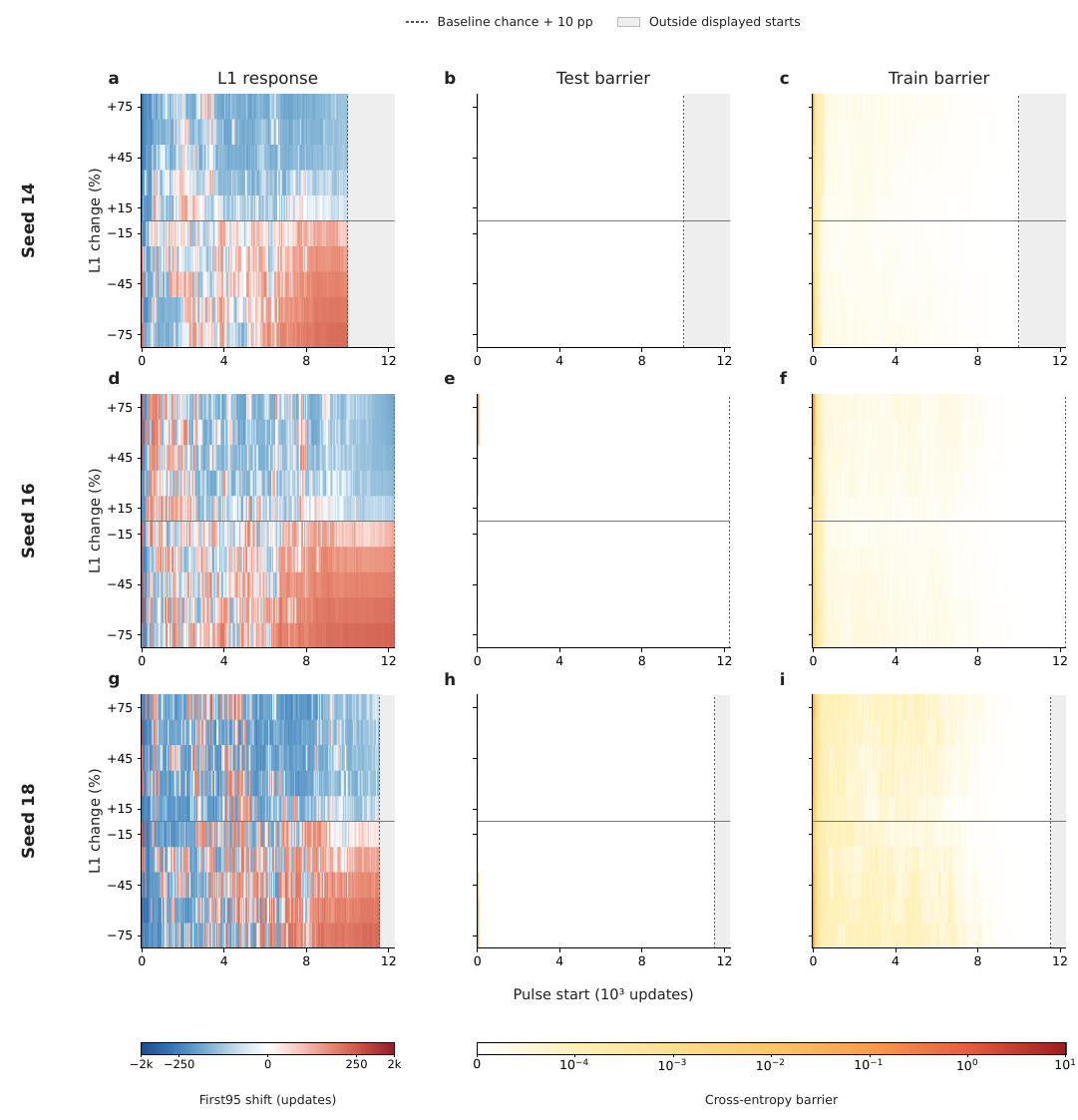}
    \caption{
    $\ell_1$ responses and loss barriers across three independent initializations.
    Rows correspond to seeds 14, 16, and 18, from top to bottom.
    Columns show generalization-time shifts, test-loss barriers, and training-loss barriers, respectively.
    Experimental settings and intervention details are provided in Appendix~\ref{app:l1}.
    }
    \label{fig:l1_response_across_seeds}
\end{figure*}
\end{document}